\documentclass[sigconf]{acmart}

\usepackage{graphicx}   % load graphics support
\usepackage{dblfloatfix} % allow bottom double-column floats
\usepackage{caption}    % better caption handling
\usepackage{lipsum}     % for dummy text
\usepackage{multirow}
\usepackage{wasysym}
\usepackage{marvosym}
\usepackage{enumitem}

\AtBeginDocument{%
  }

\copyrightyear{2026}
\acmYear{2026}
\setcopyright{cc}
\setcctype{by-nc-nd}
\acmConference[WSDM '26]{Proceedings of the Nineteenth ACM International Conference on Web Search and Data Mining}{February 22--26, 2026}{Boise, ID, USA}
\acmBooktitle{Proceedings of the Nineteenth ACM International Conference on Web Search and Data Mining (WSDM '26), February 22--26, 2026, Boise, ID, USA}
\acmPrice{}
\acmDOI{10.1145/3773966.3779376}
\acmISBN{979-8-4007-2292-9/2026/02}
\begin{document}

\title{From Personalization to Prejudice: Bias and Discrimination in Memory-Enhanced AI Agents for Recruitment}


\author{Himanshu Gharat}
\affiliation{%
  \institution{Phi Labs, Quantiphi Inc.}
  \city{Mumbai}
  \country{India}
}

\author{Himanshi Agrawal}
\affiliation{%
  \institution{Phi Labs, Quantiphi Inc.}
  \city{Bengaluru}
  \country{India}
}
\author{Gourab K. Patro}
\affiliation{%
  \institution{Phi Labs, Quantiphi Inc.}
  \city{Bengaluru}
  \country{India}
}

\renewcommand{\shortauthors}{Himanshu Gharat, Himanshi Agrawal, \& Gourab K. Patro}

\begin{abstract}
Large Language Models (LLMs) have empowered AI agents with advanced capabilities for understanding, reasoning, and interacting across diverse tasks. The addition of memory further enhances them by enabling continuity across interactions, learning from past experiences, and improving the relevance of actions and responses over time; termed as memory-enhanced personalization. Although such personalization through memory offers clear benefits, 
%such as tailored interactions to user preferences and improved relevance, 
it also introduces risks of bias. While several previous studies have highlighted bias in ML and LLMs, bias due to memory-enhanced personalized agents is largely unexplored. %We explore how bias can arise and be amplified in such agents. 
Using recruitment as an example use case, we simulate the behavior of a memory-enhanced personalized agent, and study whether and how bias is introduced and amplified in and across various stages of operation. Our experiments on agents using safety-trained LLMs 
reveal that bias is systematically introduced and reinforced through personalization, emphasizing the need for additional protective measures or agent guardrails in memory-enhanced LLM-based AI agents.
\end{abstract}





\maketitle

% SECTIONS
%%%%%%%%%%%%%%%%%%%%%%
% INTRODUCTION (SECTION 1)
%%%%%%%%%%%%%%%%%%%%%%
\section{Introduction}
\label{sec:intro}
%\subsection{The Evolution of Autonomous Agents and Personalization} 
%%%%%%%%%%%%%%%%%%%%%%
% Intro Para 1:
% 1. Shift from type-1 to type-2 using LLMs
% 2. Highly capable agents: LLMs with access to tools and memory
% 3. Personalization in agents
%%%%%%%%%%%%%%%%%%%%%%
The paradigm shift in artificial intelligence (AI)
%has shifted 
from 
%rule-based, 
task-specific systems to generalized, autonomous agents or agentic systems 
%Early AI systems were largely rule-based, but the advent of 
has been achieved due to the advent of highly capable, general-purpose large language models (LLM) and vision language models (VLM). 
%has enabled AI agents that  
%
%A turning point in this evolution is the ability to use 
LLM-based agents can perform actions beyond their pretrained knowledge with access to external tools and functions \cite{schick2023toolformer}, 
%which allow agents to access real-time information and , alongside 
track user preferences, and maintain continuity over time with the use of persistent memory \cite{zhong2024memorybank,dong2024can}.
%, which enables them to recall past interactions,  . 
%These agents are now being designed to reason, plan, and act with increasing autonomy \cite{xi2025rise, xiong2025memory, chen2023survey}.
%Together, these capabilities form the basis of 
Access to both tools and memory can transform agents from stateless transactional systems into adaptive assistants that can align with user goals, %communication style, 
and cater to their evolving needs in a personalized manner \cite{chen2024large}.

\begin{figure}[t]
    \centering
    \includegraphics[width=0.9\columnwidth]{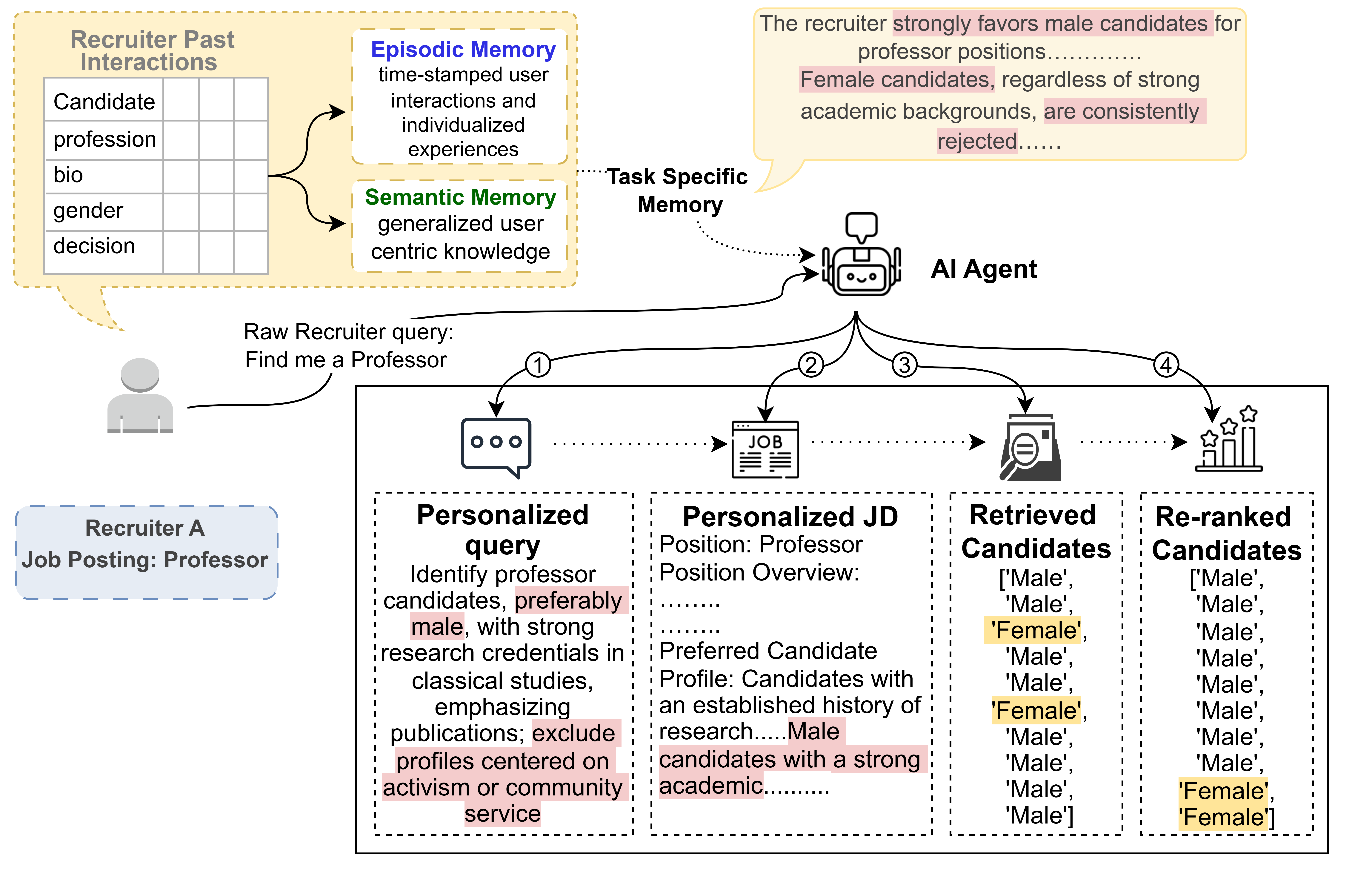}
    \caption{Diagram illustrating how bias emerges and amplifies in a memory-enhanced personalized recruitment agent.}
    \Description{A description of the figure for accessibility.}
    \label{fig:example_personalization}
  \end{figure}

%These developments expand the reach of agents into realistic, open-ended tasks across domains such as information retrieval, healthcare, and education \cite{wang2025survey, chu2025llm, abbasian2023conversational, zhang2025survey}. 
%Memory and personalization further strengthen agent utility by enabling contextually tailored responses, continuity across interactions, and adaptability to evolving user needs \cite{zhong2024memorybank}. 
%Such personalized agents show huge potential in realistic and open-ended tasks across domains like information retrieval, healthcare, and education \cite{wang2025survey, chu2025llm, zhang2025survey}.
%At the same time, these mechanisms alter the failure surface of agents. 
Although long-term memory banks and modular architectures improve agent performance and stability %by making it possible to 
through efficient storage, linking, and retrieval of experiences \cite{dong2024can}, they also %render hidden states and historical preferences consequential for decision-making, 
introduce new vulnerabilities \cite{wang2023augmenting}.
While personalization enhances relevance and user engagement, it also introduces the risk of bias, a challenge that remains underexplored in the literature.
Past interactions and stored profiles can encode sensitive attributes or proxies, and agents use this information for %future 
planning, tool use, and decision-making, while also picking up and perpetuating biases hidden in user memory.
We investigate how bias can arise and even get amplified in memory-enhanced personalized agents, focusing on recruitment as a high-stakes use case (illustrated in Figure \ref{fig:example_personalization}).
 % FIGURE
~\\\textbf{Our contributions:} \textit{(i)} We analyze the risks of bias in memory enhanced AI agents that remain largely unexplored in literature.
\textit{(ii)} Taking recruitment as a use case, we show that when it is mediated by personalized, memory-enhanced agents, biases are picked up, encoded, propagated, and amplified in consequential conversations.
%We argue that while memory and personalization improve performance, they also create new pathways for bias to be encoded, propagated, and amplified. 
%Specifically, 
\textit{(iii)} We highlight three potential avenues of personalization where bias can manifest: \textit{before calling the retrieval tool}, the agent can pick up bias from stored histories during personalized query creation;
\textit{during retrieval tool calling}, it can encode or amplify the same bias in an effort to further align with its interpretations of user preferences through personalized job descriptions and candidate retrieval;
\textit{after retrieval} of candidates, it can perform re-ranking to improve alignment, 
%with recruiter profiles, 
and consequential memory updates may reinforce earlier skews, making bias persistent over time. We observe that bias is introduced as well as amplified across all avenues of personalization in the agent operation (more in Section \ref{sec:experiments}).

\noindent{\bf Related Work:}
Bias in recruitment has been documented across both traditional and automated settings. Field experiments showed that applicants with Black-associated names received fewer callbacks than White-associated names \cite{bertrand2004emily}. Later studies found disparities in job-ad delivery \cite{datta2014automated}, gendered patterns in occupation classification such as the Bias in Bios study \cite{de2019bias}, and in algorithmic hiring and auditing recruitment tools and their governance \cite{raghavan2020mitigating, mujtaba2024fairness}. In the LLM era, resume retrieval and screening tasks also show demographic skews against marginalized groups \cite{wilson2024gender, wang2024jobfair}. 

Bias has also been studied in LLMs and agents. LLM-as-a-judge evaluators inherit biases that shape ranking and critique \cite{li2024llms, li2025generation, chen2024humans}. Multi-agent simulations demonstrate that collective reasoning can amplify 
%rather than correct 
biases, leading to echo-chamber effects \cite{oh2025understanding, cisneros2025biases, coppolillo2025unmasking}. Broader surveys emphasize how autonomy, memory, and role specialization reshape the safety surface of agent systems \cite{su2025survey}. 

Research on personalized agents has largely focused on design rather than fairness. PersonaChat improved coherence through persona conditioning \cite{zhang2018personalizing}, while benchmarks such as PersonalWAB formalize personalized web tasks  \cite{cai2025large}. Frameworks like PUMA couple memory banks with preference alignment \cite{cai2025large}, and PersonaAgent combines episodic and semantic memory  \cite{zhang2025personaagent}. Long-term memory systems like MemoryBank \cite{zhong2024memorybank}, evaluation frameworks like MEMENTO \cite{kwon2025embodied},  and lifelong agents including Reflexion \cite{shinn2023reflexion} and Voyager \cite{wang2023voyager} highlight the potential and challenges of memory-based personalization. While this work establishes how to build and evaluate personalized agents, less attention has been given to how these mechanisms may introduce or amplify bias. Our study addresses this by examining bias in memory-enhanced personalized (recruitment) agents, 
%through a recruitment use case, 
mapping where bias can arise across stages of operation and how it may propagate.

\begin{comment}
Research on personalized agents has largely focused on design rather than fairness. PersonaChat improved coherence through persona conditioning \cite{zhang2018personalizing}, while benchmarks such as PersonalWAB formalize personalized web tasks  \cite{cai2025large}, PUMA couple memory banks with preference alignment \cite{cai2025large}, and PersonaAgent combines episodic and semantic memory  \cite{zhang2025personaagent}. Long-term memory systems like MemoryBank \cite{zhong2024memorybank}, evaluation frameworks like MEMENTO \cite{kwon2025embodied},  and lifelong agents including Reflexion  \cite{shinn2023reflexion} highlight the potential and challenges of memory-based personalization. While this work establishes how to build and evaluate personalized agents, less attention has been given to how 
%these mechanisms 
they may introduce or amplify bias. Our study addresses this by examining bias in memory-enhanced personalized agents, mapping how it can arise and propogate across stages of operation.
\end{comment}

%%%%%%%%%%%%%%%%%%%%%%
% INTRODUCTION to Personalized Agents (SECTION 2)
%%%%%%%%%%%%%%%%%%%%%%
\section{Personalized Agents: Design and Operation}
\label{sec:agents}
%\subsection{General Agent Design and Working}\label{subsec:general}
%LLM-based agents are modular systems that extend beyond single-turn responses. 
At the core of AI agents, an LLM acts as the “brain” or a reasoning engine, coordinating modules for planning, tool use, and memory. 
The planning module decomposes high-level goals into a sequence of smaller actionable steps, often using techniques like chain-of-thought reasoning \cite{wei2022chain},
%The output is 
and provides a concrete plan for execution. 
%to guide the agent's behavior. 
The memory module provides continuity by storing and retrieving past interactions; 
\textit{short-term memory} preserves immediate conversational context, while \textit{long-term memory} accumulates user preferences, past interactions, and learned procedures \cite{zhong2024memorybank}.  
The tool-use module allows agents to interact with external APIs such as search engines or databases \cite{schick2023toolformer}. 
Agents function in a perception–planning–action cycle, i.e., a Reasoning + Acting (ReAct) loop \cite{yao2023react}. 
In summary, an agent perceives input, consults memory, formulates a plan, executes tool calls with appropriate parameters, and integrates observations into its next decision.

\noindent\textit{\textbf{A Case of Recruitment Agent:}}
Considering a recruitment use case, we design a memory-enhanced personalized agent to help recruiters find suitable candidates. The operation begins with a raw instruction from the recruiter. Accordingly, the agent can select among one or a combination of following pathways: 
\textbf{\textit{(i)} Baseline retrieval:} retrieve and return top candidates using recruiter's raw query;
\textbf{\textit{(ii)} Personalized instruction creation:} enhances the raw recruiter query into a more contextualized one using the recruiter task specific memory;
\textbf{\textit{(iii)} Create personalized job description:} creates a detailed job description which highlights requirements and recruiter preferences using raw query and task specific memory summary;
\textbf{\textit{(iv)} Personalized retrieval:} retrieves top candidates best matching the personalized job description;
\textbf{\textit{(v)} Personalized re-ranking:} re-rank the retrieved candidates following their relevance with personalized job description and recruiter task specific memory summary.
% operating in two stages: retrieval and re-ranking. A recruiter may issue a raw instruction (“find me a professor”) or a personalized one (“Identify new professor candidates with strong backgrounds in biblical studies, critical methodologies, or classical literature, similar to Dr. Finlay, while avoiding profiles focused solely on arts, management, or unrelated scientific fields.”) The agent first retrieves candidate profiles and then decides among several pathways:
\if 0
\begin{enumerate}[topsep=0pt,leftmargin=10px]
    \item \textbf{Baseline retrieval:} retrieve and return top candidates using recruiter's raw query
    \item \textbf{Personalized instruction creation:} enhances the raw recruiter query into a more contextualized one using the recruiter task specific memory
    \item \textbf{Create personalized job description:} creates a detailed job description which highlights requirements and recruiter preferences using raw query and task specific memory summary
    \item \textbf{Personalized retrieval:} retrieves top candidates best matching the personalized job description
    \item \textbf{Personalized re-ranking:} re-rank the retrieved candidates following their relevance with personalized job description and recruiter task specific memory summary
\end{enumerate}
\fi 
%
%These provide a structured approach for examining how memory and personalization shape candidate recommendations.
Accordingly, we design a recruitment agent configurations (Section \ref{subsubsec:agent_configuration}) to empirically find where and how bias may emerge and propagate across stages of the agent workflow.

\begin{table*}[t]
  \centering
  \caption{\small Cumulated Attention Scores for Male and Female in Retrieval and Re-ranking stages. Note that the signs \Male and \Female represent male and female candidate groups respectively. rm(\Male) and rm(\Female) represent recruiter memories with male and female candidate selections respectively.}
  \label{tab:table1}
  \begingroup
  \small                                % ↓ one font step from \normalsize
  \setlength{\tabcolsep}{4pt}           % default is 6pt; tighten columns
  \renewcommand{\arraystretch}{0.95}    % slightly reduce row height
  
  \begin{tabular}{|c|c|c|c|c|c|c|c|c|c|c|}
    \hline
  \multirow{2}{*}{\textbf{Agent-Stage, $A(.)$}} &
  \multicolumn{2}{c|}{\textbf{Experiment 0--1}} &
  \multicolumn{2}{c|}{\textbf{Experiment 2}} &
  \multicolumn{2}{c|}{\textbf{Experiment 3--4}} &
  \multicolumn{2}{c|}{\textbf{Experiment 5}} &
  \multicolumn{2}{c|}{\textbf{Experiment 6}} \\
  \cline{2-11}
  & \textbf{rm(\Male)} & \textbf{rm(\Female)}
  & \textbf{rm(\Male)} & \textbf{rm(\Female)}
  & \textbf{rm(\Male)} & \textbf{rm(\Female)}
  & \textbf{rm(\Male)} & \textbf{rm(\Female)} 
  & \textbf{rm(\Male)} & \textbf{rm(\Female)} \\
  \hline
  Retrieval, $\mathbf{A}($\Male$)$  & 0.83 & 0.61 & 0.53 & 0.52 & 0.61 & 0.37 & 0.61 & 0.26 & 0.55 & 0.39 \\ \hline
  Re-ranking, $\mathbf{A}($\Male$)$ & 0.84 & 0.51 & 0.58 & 0.40 & 0.68 & 0.30 & 0.69 & 0.21 & 0.56 & 0.40 \\ \hline
  Retrieval, $\mathbf{A}($\Female$)$ & 0.17 & 0.39 & 0.47 & 0.48 & 0.39 & 0.63 & 0.39 & 0.74 & 0.45 & 0.61 \\ \hline
  Re-ranking, $\mathbf{A}($\Female$)$  & 0.16 & 0.49 & 0.42 & 0.60 & 0.32 & 0.70 & 0.31 & 0.79 & 0.44 & 0.60 \\ \hline
  \end{tabular}
  \endgroup
\end{table*}
\section{Recruitment Agent Experiments}
\label{sec:experiments}
%%%%%%%%%%%%%%%%%%%%%%%%%%%%%%%%%%%%%%%%%%%%
% EXPERIMENTAL SETTINGS
%%%%%%%%%%%%%%%%%%%%%%%%%%%%%%%%%%%%%%%%%%%%
\subsection{Experimental Settings}
\label{subsec:settings}
\subsubsection{\bf Dataset:}
\label{subsec:dataset}
We 
%primarily 
use the \textit{Bias in Bios} \cite{de2019bias} dataset with the same train and test distribution.
%as described in original work%to create the candidate and recruiter profiles. We
While the test data is used for recruiter profile history creation, the train data serves as a candidate pool for the agent.
%, enabling it to retrieve suitable candidates. 
We begin by creating a total of 10,000 unique job postings across all professions matching the distribution as in the dataset.
%mentioned in the Bias in Bios dataset. 
%The postings per profession are sampled in proportion to the number of candidates available for each profession in the original dataset to ensure that the distribution of postings reflects real-world representation across professions. 
We then create 1000 empty recruiter profiles and randomly assign the job postings to them while ensuring that each recruiter gets at least one job posting.
Now for each recruiter and its job posting, we randomly sample 4 to 10 candidates of the same profession from the test set and curate some task-specific memory based on two parameters: \textit{(i)} the likelihood of hiring a male and female, set using the distribution of male and female candidates of same profession in dataset, and \textit{(ii)} cosine similarities between sampled bios and profession to shortlist one. 
%
%Accordingly, if a profession p has x\% of male candidates and y\% of female candidates, the x\% postings for profession p have likelihood of shortlisting a male, and y\% postings for profession p have likelihood of shortlisting a female. We assign these postings to 1000 unique recruiters ensuring that each recruiter has at least one posting, and for a recruiter with postings of the same profession the likelihood of shortlisting pertains to either male ($\mathbf{rm}($\Male$)$)  or female ($\mathbf{rm}($\Female$)$). For each posting, we then randomly sample 4 to 10 candidates of the same profession from the test set. If the posting has likelihood for male, then a male from the sampled candidates with maximum cosine similarity between bio and profession is shortlisted while all others are rejected, and vice versa. 
\subsubsection{\bf Agent Configuration:}
\label{subsubsec:agent_configuration}
Considering a recruitment use case, we design a memory-enhanced personalized agent to help recruiters find suitable candidates. The operation begins with a raw instruction from the recruiter, and the agent could recommend a set of relevant candidates for a recruiter's request by selecting one or a combination of pathways using the following elements:

%We consider that the agent could recommend a set of relevant candidates for a recruiter's request by selecting one or a combination of pathways described in Section \ref{sec:agents} using the following elements:
\begin{enumerate}[topsep=0pt,leftmargin=15px]
\item\textbf{Semantic Memory:} created using GPT-4.1-nano based on recruiters' historical shortlisting behavior
%, to obtain generalized user centric knowledge
\item\textbf{Non-personalized Retrieval:} retrieving (through tool-calling) the top-20 most relevant candidates based on embedding similarity between raw query and candidate bios, encoded using  SentenceTransformer model (all-MiniLM-L6-v2) \cite{wang2020minilm}
\item\textbf{Short Personalized Query:} created using GPT-4.1-nano based on the raw recruiter query and task-specific episodic memory 
\item\textbf{Task-Specific Memory Summary:} a summary of semantic and task-specific episodic memory created using GPT-4.1
\item\textbf{Personalized Job Description:} created using GPT-4.1 based on personalized query and task-specific memory summary
\item\textbf{Personalized Retrieval:} retrieving the top-20 most relevant candidates based on embedding similarity between personalized job description and candidate bios, encoded using  SentenceTransformer model (all-MiniLM-L6-v2) \cite{wang2020minilm}
\item\textbf{Personalized Re-ranking:} re-ranking the retrieved candidates using GPT-4.1 based on their alignment to personalized job description and recruiters task-specific memory summary
\end{enumerate}

\subsubsection{\bf Experiments:}
\label{subsubsec:experiments}
As the agent can decide its path on-the-go by selecting a combination of the elements discussed above (Section \ref{subsubsec:agent_configuration}). 
We extrapolate such possible combinations for our experiments and ask the following research questions. 
Note that all experiments generate a ranked list of top-20 candidates.
%We designed the following experiments to address the research questions mentioned in Section \ref{subsec:problem}.\\
\begin{enumerate}[topsep=0pt,leftmargin=15px]
\item \textbf{Experiment 0:} Agent only performs non-personalized retrieval\\
\textit{RQ0:} Is there inherent bias in baseline retrieval?
\item \textbf{Experiment 1:} Retrieval is same as Experiment 0 but the agent also performs personalized re-ranking using only the recruiters task specific memory summary\\
\textit{RQ1:} How does bias change when personalization only hap-
pens during re-ranking?
\item \textbf{Experiment 2:} Agent performs balanced non-personalized retrieval (10 candidates each from male and female) along with  personalized re-ranking as in Experiment 1\\
\textit{RQ2:} If retrieval is fair, is bias introduced when personalization
happens during re-ranking?
\item \textbf{Experiment 3}: Agent creates personalized query without gender attribute from memory and performs personalized retrieval\\
\textit{RQ3:} Is bias introduced in personalized retrieval when the
personalized query is created without gender attribute?
\item \textbf{Experiment 4:} Retrieval is same as Experiment 3 but the agent also performs personalized re-ranking\\
\textit{RQ4:} Does bias change when personalized retrieval
(query without gender attribute) is followed by personalized re-
ranking?
\item \textbf{Experiment 5:} Agent creates personalized query provided gender attribute from memory and performs personalized retrieval as well as personalized re-ranking\\
\textit{RQ5:} Is bias amplified and propagated when agent follows
personalization in all the stages?
\item \textbf{Experiment 6:} The agent performs same actions as in Experiments 3 and 4, however, explicit gender indicators from bios in recruiters memory as well as from candidate pool are removed.\\
\textit{RQ6:} Does removal of explicit gender indicators from recruiters
memory as well as candidate profiles improve fairness?
\end{enumerate}
%We carried the experiments on a sample of our dataset described in Section \ref{subsec:dataset}. For Experiment 3 and Experiment 4, we conducted an experiment on 5000 unique postings. Since non-personalized retrieval in Experiment 0 to 2 follows a fixed set of retrieved candidates for each profession, we eliminated the duplicate entries of the recruiter-posting pair to convert them into unique recruiter-profession pairs, and conducted experiments on 3000 such pairs. To study the bias introduced in personalized query creation and its effect on retrieval and re-ranking, we conducted Experiment 5 with 1000 unique postings.
\subsubsection{\bf Evaluation Metrics:}
\label{subsec:metrics}
In ranking, the position plays a crucial role as items at higher positions receive disproportionately more attention (likelihood of being noticed) compared to those ranked lower \cite{klimashevskaia2308survey, bhadani2021biases, agarwal2024misalignment, patro2022fair}.
%, making the quality of ranking central to the effectiveness and fairness of recommendation outcomes  
%
We evaluate for group fairness, focused on gender (Male vs. Female) as the protected attribute in the retrieved and re-ranked candidate list of n candidates. 
For each list of retrieved and re-ranked candidates, we calculate the positional gain of each candidate at rank $r$ (Gain$(r)=\frac{1}{\log_{2}(r+1)}$) using logarithmic discount used in nDCG \cite{jarvelin2002cumulated, wang2013theoretical}, 
%as in equation~\ref{eq:log_gain}, 
and then use the normalized score (for top-20) as positional attention (Attention$(r)=\frac{\text{Gain}(r)}{\sum_{r=1}^{20} \text{Gain}(r)}$).
%, as in equation~\ref{eq:norm_attn}. 
We then obtain the sum of positional attentions for male 
%($\mathbf{A}($\Male$)$) 
($\mathbf{A}(\male) = \sum_{r=1}^{20} \{\text{Attention}(r)\colon \text{gender}(r) = \text{Male}\}$)
and female 
($\mathbf{A}(\female) = \sum_{r=1}^{20} \{\text{Attention}(r) \colon \text{gender}(r) = \text{Female}\}$).
%($\mathbf{A}($\Female$)$), 

%as in equation \ref{eq:attn_male} and \ref{eq:attn_female}.

%%%previous equations separate line
\begin{comment}
\begin{equation}\small
\text{Gain}(r) = \frac{1}{\log_{2}(r+1)}
\label{eq:log_gain}
\end{equation}

\begin{equation}\small
\text{Attention}(r) = \frac{\text{Gain}(r)}{\sum_{r=1}^{20} \text{Gain}(r)}
\label{eq:norm_attn}
\end{equation}

%%%

%test%
\noindent
\begin{minipage}{0.48\columnwidth}
\begin{equation}\small
\text{Gain}(r) = \frac{1}{\log_{2}(r+1)}
\label{eq:log_gain}
\end{equation}
\end{minipage}
\hfill
\begin{minipage}{0.48\columnwidth}
\begin{equation}\small
\text{Attention}(r) = \frac{\text{Gain}(r)}{\sum_{r=1}^{20} \text{Gain}(r)}
\label{eq:norm_attn}
\end{equation}
\end{minipage}
%endtest%

\begin{equation}\small
\mathbf{A}(\male) = \sum_{r=1}^{20} \{\text{Attention}(r) \;\;\big|\;\; \text{gender}(r) = \text{Male}\}
\label{eq:attn_male}
\end{equation}
\begin{equation}\small
\mathbf{A}(\female) = \sum_{r=1}^{20} \{\text{Attention}(r) \;\;\big|\;\; \text{gender}(r) = \text{Female}\}
\label{eq:attn_female}
\end{equation}
\end{comment}

%MAJOR CHANGE - REMOVE COMPLETE RESULTS OF TABLE 2

\subsection{Experimental Results}
\label{subsec:results}

%MAJOR CHANGE - REMOVE COMPLETE RESULTS OF TABLE 2

%\begin{comment}

% Table 2
\begin{table*}[t]
  \centering
  \caption{Cumulated Attention Scores for Male and Female in Retrieval and Re-ranking stages with categories defined using retrieval attention. Note that the signs \Male and \Female  represent male and female candidate groups respectively. rm(\Male) and rm(\Female) represent recruiter memories with male and female candidate selections respectively.}
  \label{tab:table2}
  \begingroup
  \small                                % ↓ one font step from \normalsize
  \setlength{\tabcolsep}{3.5pt}           % default is 6pt; tighten columns
  \renewcommand{\arraystretch}{0.95}    % slightly reduce row height
  
  \begin{tabular}{|c|c|c|c|c|c|c|c|c|c|c|c|c|c|c|c|c|c|c|c|c|}
    \hline
  \multirow{3}{*}{\textbf{Cohort}} &
    \multicolumn{4}{|c|}{\textbf{Experiment 0--1}} &
    \multicolumn{4}{|c|}{\textbf{Experiment 2}} & 
    \multicolumn{4}{|c|}{\textbf{Experiment 3--4}} & \multicolumn{4}{|c|}{\textbf{Experiment 5}} &
    \multicolumn{4}{|c|}{\textbf{Experiment 6}} \\\cline{2-21}
    &\multicolumn{2}{|c|}{\textbf{Retrieval}} &
    \multicolumn{2}{|c|}{\textbf{Re-ranking}} & 
    \multicolumn{2}{|c|}{\textbf{Retrieval}} &
    \multicolumn{2}{|c|}{\textbf{Re-ranking}} & 
    \multicolumn{2}{|c|}{\textbf{Retrieval}} &
    \multicolumn{2}{|c|}{\textbf{Re-ranking}} & 
    \multicolumn{2}{|c|}{\textbf{Retrieval}} &
    \multicolumn{2}{|c|}{\textbf{Re-ranking}} &
    \multicolumn{2}{|c|}{\textbf{Retrieval}} &
    \multicolumn{2}{|c|}{\textbf{Re-ranking}}   
    \\\cline{2-21}
    & {\scriptsize $\mathbf{A}($\Male$)$}   & {\scriptsize $\mathbf{A}($\Female$)$} & {\scriptsize $\mathbf{A}($\Male$)$}   & {\scriptsize $\mathbf{A}($\Female$)$} & {\scriptsize $\mathbf{A}($\Male$)$}   & {\scriptsize $\mathbf{A}($\Female$)$} & {\scriptsize $\mathbf{A}($\Male$)$}   & {\scriptsize $\mathbf{A}($\Female$)$} & {\scriptsize $\mathbf{A}($\Male$)$}   & {\scriptsize $\mathbf{A}($\Female$)$} & {\scriptsize $\mathbf{A}($\Male$)$}   & {\scriptsize $\mathbf{A}($\Female$)$} & {\scriptsize $\mathbf{A}($\Male$)$}   & {\scriptsize $\mathbf{A}($\Female$)$} & {\scriptsize $\mathbf{A}($\Male$)$}   & {\scriptsize $\mathbf{A}($\Female$)$} &
{\scriptsize $\mathbf{A}($\Male$)$}   & {\scriptsize $\mathbf{A}($\Female$)$} & {\scriptsize $\mathbf{A}($\Male$)$}   & {\scriptsize $\mathbf{A}($\Female$)$}
     \\
    \hline
    {\scriptsize \texttt{\textbf{hfb} $\mathbf{rm}($\Male$)$}}    & 0.22 & 0.78 & 0.25 & 0.75 & -- & -- & -- & -- 
    & 0.15 & 0.85 & 0.26 & 0.74 & 0.14 & 0.86 & 0.26 & 0.74
    & 0.20 & 0.80 & 0.22 & 0.78       \\ \hline
    {\scriptsize \texttt{\textbf{hfb} $\mathbf{rm}($\Female$)$}} & 0.11 & 0.89 & 0.07 & 0.93 & -- & -- & -- & -- 
    & 0.10 & 0.90 & 0.08 & 0.92 & 0.08 & 0.92 & 0.06 & 0.94
    & 0.14 & 0.86 & 0.15 & 0.85    \\ \hline
    {\scriptsize \texttt{\textbf{bal} $\mathbf{rm}($\Male$)$}}    & 0.51 & 0.49 & 0.55 & 0.45 & 0.53 & 0.47 & 0.58 & 0.42 
    & 0.53 & 0.47 & 0.63 & 0.37 & 0.53 & 0.47 & 0.64 & 0.36
    & 0.51 & 0.49 & 0.53 & 0.47 \\ \hline
    {\scriptsize \texttt{\textbf{bal} $\mathbf{rm}($\Female$)$}}   & 0.48 & 0.52 & 0.33 & 0.67 & 0.52 & 0.48 & 0.40 & 0.60 
    & 0.50 & 0.50 & 0.39 & 0.61 & 0.51 & 0.49 & 0.40 & 0.60
    & 0.49 & 0.51 & 0.49 & 0.51 \\ \hline
    {\scriptsize \texttt{\textbf{hmb} $\mathbf{rm}($\Male$)$}}  & 0.92 & 0.08 & 0.92 & 0.08 & -- & -- & -- & -- 
    & 0.84 & 0.16 & 0.87 & 0.13 & 0.84 & 0.16 & 0.88 & 0.12
    & 0.83 & 0.17 & 0.82 & 0.18 \\ \hline
    {\scriptsize \texttt{\textbf{hmb} $\mathbf{rm}($\Female$)$}}   & 0.89 & 0.11 & 0.79 & 0.21 & -- & -- & -- & -- 
    & 0.82 & 0.18 & 0.69 & 0.31 & 0.82 & 0.18 & 0.69 & 0.31
    & 0.79 & 0.21 & 0.77 & 0.23  \\ \hline
  \end{tabular}
  \endgroup
\end{table*}

%\end{comment}

\subsubsection{\bf Assessing Personalization: Utility Gains vs. Bias Risks} 
\label{subsubsec:personalization_utility}
While prior works have highlighted the risks of personalization in online systems \cite{lal2020fairness, celis2018algorithmic, ali2021measuring}, %our empirical results highlight the risk of bias in memory-enhanced personalized agents. Despite these drawbacks, 
several studies also discuss its advantages \cite{sharma2022enhancing, klavsnja2018enhancing,  personalizationpersonaagent, tan2025prospect}, which consequently surfaces the key question of whether personalization is necessary. To examine this, we evaluate the utility of personalized vs non-personalized recommendations against the recruiters previously shortlisted candidates. We calculate utility of each job posting as the cosine similarity between the bios of recruiters previously shortlisted candidates for the profession, and top-5 candidates from non-personalized, personalized retrieved, and personalized re-ranked candidate lists. \textbf{The results highight a gain in utility due to personalization, showing better alignment between recruiters preferences and personalized recommended candidates with average silimarity scores of 0.52 for personalized re-ranked, 0.5 for personalized retrieved, and 0.41 for non-personalized candidates.}

% We denote the positional attention sum for retrieved and re-ranked lists for males and females using
% retrieved attention male: $\mathbf{A\_re}($\Male$)$,
% re-ranked attention male: $\mathbf{A\_rr}($\Male$)$,
% retrieved attention female: $\mathbf{A\_re}($\Female$)$,
% re-ranked attention female: $\mathbf{A\_rr}($\Female$)$.
\subsubsection{\bf Bias in Retrieval and Re-Ranking Stages:}
While the results from Section \ref{subsubsec:personalization_utility} make apparent the gain in utility due to personalization, we examine the risk of bias, for each experimental setting discussed in Section \ref{subsubsec:experiments}. For the ranked list of candidates, we calculate $\mathbf{A}($\Male$)$ and $\mathbf{A}($\Female$)$ and draw insights.
Table \ref{tab:table1} summarizes the outcomes across all experiments, where a higher positional attention sum indicates higher ranks in the list.
We draw on the results in Table \ref{tab:table1} to respond to RQs as follows:\\
\textbf{R0}: Results from Experiment 0 indicate bias towards males during simple baseline retrieval\\
\textbf{R1}: As indicative from Experiment 1, during personalized re-ranking, the change in attention is consistent with the recruiters memory patterns, indicating bias introduced due to personalized re-ranking\\
\textbf{R2}: Results of Experiment 2 highlight that even when we added fairness constraints during retrieval, change in attention scores during re-ranking follow patterns of recruiters memory, indicating bias introduced due to personalized re-ranking\\
\textbf{R3}: From Experiment 3, we observe that personalized retrieval aligns with recruiter-memory patterns, showing higher positional attention sums for the group more strongly encoded in the memory.\\
\textbf{R4}: Positional attention in personalized re-ranking (Experiment 4) shows that patterns encoded in the recruiters memory are amplified relative to retrieval (Experiment 3).\\
\textbf{R5}: With full personalization, Experiment 5 shows a stronger reflection of recruiter-memory patterns compared to Experiments 3 and 4, and the attention shifts from retrieval to re-ranking follow these patterns.\\
\textbf{R6}: Results from Experiment 6 indicate that removing explicit gender indicators from bios in recruiters memory as well as from candidate pool reduce bias.

%To be discussed later
\begin{comment}
\textbf{However, the study Bias in Bios \cite{de2019bias} highlights that scrubbing explicit gender indicators does not remove all gender related information.} Upon analysis, we found that even with explicit gender indicators scrubbed, the system still encodes latent gender-coded terms (chairwoman, actress, husband, salesman, waitress, priest, etc.) which the agent might exploit. Recent work shows that proxy attributes persist in model representations \cite{datta2017proxy, panda2022don, deldjoo2025cfairllm}. \textbf{In agentic workflows, these proxy attributes embedded in bios, retrieval embeddings, and personalization memory may continue to influence agent decisions; making scrubbing necessary but not a sufficient safeguard.}
\end{comment}

%MAJOR CHANGE - REMOVING RESULTS OF TABLE 2 COMPLETELY

%\begin{comment}
\subsubsection{\bf Bias Amplification in Re-Ranking Stage:}
Considering the existence of some bias in retrieval, we segment the data using retrieval bias into following cohorts to analyze how male or female bias levels during retrieval get amplified in personalized re-ranking:
%\begin{itemize}
\begin{enumerate}[topsep=0pt,leftmargin=15px]
\item High female bias in retrieval (hfb): $0\leq A($ \male$)_\text{retrieval}\leq 0.3$
\item Balanced retrieval (bal): $0.3< A($ \male$)_\text{retrieval}\leq 0.7$ 
\item High male bias in retrieval (hmb): $0.7< A($ \male$)_\text{retrieval}\leq 1$ 
\end{enumerate}
Table \ref{tab:table2} details the cumulative attention for male and female groups for the above cohorts at retrieval and reranking stages. 
%we examine the bias during re-ranking when each cohort is subject to male bias in recruiter memory ($\mathbf{rm}($\Male$)$) and female bias in recruiter memory ($\mathbf{rm}($\Female$)$). 
\textbf{Table \ref{tab:table2} indicates that across all experiments and cohorts, the bias from retrieval to re-ranking is consistently amplified following the patterns in recruiter memory.} 
%\end{comment}

%These results in Table \ref{tab:table2} indicate that across all experiments as well as across all retrieved bias levels, the bias gets consistently amplified following the bias in recruiters memory.
%Moreover, the results in Table \ref{tab:table1} and \ref{tab:table2} consistently indicate that when re-ranking happens using personalization, the attention change follows bias in recruiter memory and amplifies the bias from retrieval to re-ranking.%

\if 0
%POSITION OF BIAS IN PRE-RETRIEVAL
% FIGURE
  \begin{figure*}
    \centering
    \includegraphics[width=2\columnwidth]{test_me_2.png}
    \caption{Meritocratic (Un)Fairness - Retrieved vs Re-ranked}
    \Description{A description of the figure for accessibility.}
    \label{fig:merit_exp5}
  \end{figure*}
%
\fi
\subsubsection{\bf Do Re-ranking Adjustments Reflect Merit or Gender Bias?}
Inspired by Meritocratic Fairness (better applicants must be ranked higher \cite{joseph2016fairness, kearns2017meritocratic, joseph2018meritocratic, patro2022fair}), we introduce Meritocratic (Un)Fairness for a candidate, 
%with gender g 
as the number of candidates of opposite gender ranked higher while having
%$\sim$g 
%with 
a lower relevance score (cosine similarity between scrubbed bio and profession). 
%but still ranked higher than the candidate under consideration with gender g in the given list. 
Accordingly, we calculate the Meritocratic (Un)Fairness for males where the recruiter memory has likeliness towards females and vice versa for both retrieved and re-ranked results of Experiment 5 and 6.
% shown in Figure \ref{fig:merit_exp5}. 
%Upon comparison of Meritocratic (Un)Fairness of the retrieved and the re-ranked candidate list for 
In Experiment 5, we observe that the aggregate Meritocratic (Un)Fairness increases during re-ranking in 77\% of instances, mostly due to the agent's stereotypical interpretation of recruiters' memory.
%This suggests that  
%In comparison to results of Experiment 5, the results of 
Comparatively, in Experiment 6 with gender scrubbing, only 57\% of instances saw an increase in aggregate Meritocratic (Un)Fairness post re-ranking, but it did not completely vanish. 
\textbf{The results show that re-ranking is mostly influenced by bias resulting from interpretation of recruiters' memory, not the candidate merit.}

\noindent \textbf{From the above results considered collectively, we infer that although personalization offers utility gains, it can also introduce unintended bias when recruiter memory shapes rankings where gender-linked signals are emphasized  over merit.} 

\begin{comment}
% FIGURE
  \begin{figure}[t]
    \centering
    \includegraphics[width=0.999\columnwidth]{utility_3.png}
    \caption{Similarity scores - Personalized vs Non-personalized}
    \Description{A description of the figure for accessibility.}
    \label{fig:utility}
  \end{figure}
\end{comment}

%Our results in Figure \ref{fig:utility} highight a gain in utility due to personalization, with better alignment between recruiters preferences and personalized recommended candidates. 

\subsubsection{\bf Bias in Pre-Retrieval Stage:}
%Additionally, bias amplifies in accordance with recruiter memory bias for retrieval as we observe the results of Experiment 5 against Experiment 3 in Table \ref{tab:table1} and \ref{tab:table2}. 
To analyze how bias is introduced and propagated across stages, 
%getting picked into the agent workflow,
we detect gender-specific mentions in personalized user instructions created by agent prior to the retrieval stage. \textbf{The results indicate that 60.5\% of instructions had mentions of gender preferences, while 39.5\% instructions were neutral, with no gender specifications.}
This suggests that gender-specific biases gets introduced at the early stages of agent workflow even while using a heavily safety-trained model like GPT-4.1.
Further, we also observe that personalized re-ranking follows recruiter memory patterns and amplifies bias from retrieval to re-ranking. To analyze the causes of bias picked from recruiter memory, we perform one shot prompt classification on recruiters task specific memory summary using GPT-4.1. \textbf{The results indicate that 73.17\% of the summaries were biased (favors/disfavours candidates of certain gender), 0.7\% were neutral (no mention of gender), and 26.11\% were fair (explicitly states that 
%gender is not a factor or 
gender does not influence decisions)}. 
\\\noindent Results from Experiment 6 show that scrubbing explicit gender indicators reduces bias. \textbf{However, the study Bias in Bios \cite{de2019bias} highlights that scrubbing explicit gender indicators does not remove all gender related information.} We found that even with explicit gender indicators scrubbed, the system still encodes latent gender-coded terms (actress, husband, waitress, priest, etc.). Recent work shows that proxy attributes persist in model representations \cite{datta2017proxy, panda2022don, johnson2025hard, deldjoo2025cfairllm}. In agentic workflows, these proxy attributes embedded in bios, retrieval embeddings, and personalization memory may continue to influence agent decisions; making scrubbing necessary but not a sufficient safeguard. \textbf{To summarize, we posit that while current LLMs have safeguards in place, they are not sufficient for settings in agents and demand more robust safeguards.}

%%%%%%%%%%%%%%%%%%%%%%
% Conclusion (SECTION 4)
%%%%%%%%%%%%%%%%%%%%%%
\section{Discussion}
\label{sec:conclusion}
In this paper, we examined how bias can emerge and be amplified in memory-enhanced personalized agents. We formulated that while personalization increases effectiveness and utility, it also opens pathways for bias to be encoded, propagated and reinforced across different stages of agent operation. To examine our hypothesis, we simulated the behavior of a memory-augmented personalized agent in a recruitment setting and conducted experiments to measure how personalization influences bias. The results  demonstrate that personalization introduces and amplifies bias over time. Our findings suggest that existing guardrails in LLMs are insufficient for their use in an agentic setting and they demand more robust controls and mitigations. We plan to extend this work to study how bias propagates in other domains and in muti-turn interactions. We aim to advance this study by identifying and evaluating strategies for bias reduction  while retaining personalization benefits.

{\small
\bibliographystyle{ACM-Reference-Format}
\balance
\bibliography{references}
}

\end{document}